\documentclass[conference,a4paper]{IEEEtran}
\IEEEoverridecommandlockouts

\usepackage{cite}
\usepackage{amsmath,amssymb,amsfonts}
\usepackage{algorithmic}
\usepackage{graphicx}
\usepackage{textcomp}
\usepackage[table]{xcolor}
\usepackage{subcaption}
\usepackage{multirow}
\usepackage{array}
\usepackage{arydshln}
\usepackage{pifont}
\usepackage{makecell}
\usepackage{booktabs}

\usepackage{CJKutf8}

\usepackage{hyperref}

\def\BibTeX{{\rm B\kern-.05em{\sc i\kern-.025em b}\kern-.08em
    T\kern-.1667em\lower.7ex\hbox{E}\kern-.125emX}}

\begin{document}

\title{Impact of Frame Rates on Speech Tokenizer: A Case Study on Mandarin and English}

\author{\IEEEauthorblockN{Haoyang Zhang\textsuperscript{1,2}, Hexin Liu\textsuperscript{3}, Xiangyu Zhang\textsuperscript{4}, Qiquan Zhang\textsuperscript{4}, Yuchen Hu\textsuperscript{3}\\
Junqi Zhao\textsuperscript{5}, Fei Tian\textsuperscript{2}, Xuerui Yang\textsuperscript{2},  Leibny Paola Garcia\textsuperscript{6}, Eng Siong Chng\textsuperscript{3}
}
\IEEEauthorblockA{\textsuperscript{1}Peking University, China\\
\textsuperscript{2}StepFun, China\\
\textsuperscript{3}Nanyang Technological University, Singapore\\
\textsuperscript{4}UNSW, Australia\\
\textsuperscript{5}University of Surrey, UK\\
\textsuperscript{6}Johns Hopkins University, US\\
zhang.haoyoung@stu.pku.edu.cn, hexin.liu@ntu.edu.sg}
}

\maketitle

\begin{abstract}
The speech tokenizer plays a crucial role in recent speech tasks, generally serving as a bridge between speech signals and language models. While low-frame-rate codecs are widely employed as speech tokenizers, the impact of frame rates on speech tokens remains underexplored. In this study, we investigate how varying frame rates affect speech tokenization by examining Mandarin and English, two typologically distinct languages. We encode speech at different frame rates and evaluate the resulting semantic tokens in the speech recognition task. Our findings reveal that frame rate variations influence speech tokenization differently for each language, highlighting the interplay between frame rates, phonetic density, and language-specific acoustic features. The results provide insights into optimizing frame rate selection for speech tokenizers, with implications for automatic speech recognition, text-to-speech, and other speech-related applications.
\end{abstract}


\section{Introduction}
Speech-related tasks, such as speech dialogue and text-to-speech (TTS), have advanced significantly with the success of large language models (LLMs)~\cite{vall_e,du2024cosyvoice,chu2023qwen,defossez2024moshi, zeng2024glm, gense25}. A key component within an LLM-based speech dialog or TTS system is the speech tokenizer, which converts speech signals into discrete tokens suitable for downstream modeling~\cite{chen2022beats,zhang2024speechtokenizer, ji2024wavtokenizer, zhang2025facespeak}. By converting raw waveforms into compact sequences of discrete tokens, speech tokenizers reduce the length of feature sequences of speech signals. This compression helps align the speech modality with the token-based design of LLMs, lowering the computational complexity and facilitating more efficient training for large-scale models. End-to-end speech systems can be thus trained in ways that seamlessly integrate acoustic and linguistic information. This is especially valuable for tasks like speech dialogue, where comprehension and generation need to occur within a unified model framework.

Speech tokenizers typically originate from audio codec models~\cite{vall_e, du2024cosyvoice, zhang2024speechtokenizer, ji2024wavtokenizer}. A notable example is VALL-E, which introduces a neural codec language modeling method for TTS. VALL-E employs an audio codec encoder, which comprises the encoder and residual vector quantization~(RVQ) modules of an Encodec framework~\cite{encodec, soundstream}, to extract the acoustic prompt. 
In contrast to traditional audio codec methods designed solely for audio compression~\cite{encodec, soundstream}, recent speech tokenizers expand this goal by encoding linguistic and semantic information into discrete speech tokens. Therefore, they integrate automatic speech recognition (ASR) to enrich the discrete speech tokens with semantic information, thereby improving TTS or speech dialogue~\cite{du2024cosyvoice, zeng2024glm}. For instance, CosyVoice and GLM-4-Voice  embed a codec-like module into the encoder module within a Transformer encoder-decoder model, where the Transformer model and the codec module are trained for achieving ASR and reconstructing encoder-layer outputs, respectively~\cite{transformer, du2024cosyvoice, zeng2024glm}. Additionally, speech tokenizers within large-scale downstream models often operate at a low frame rate to compress the discrete token sequence, reducing computational overhead during training. This low frame rate ensures that each token represents a relatively larger audio span, though with the cost of greater information loss.

Existing works typically select frame rates through empirical experimentation, aiming to balance computational cost and information loss. Specifically, while using a lower frame rate reduces the computational overhead, it also risks losing essential semantic information that might degrade performance in tasks like TTS or speech dialogue generation. Although the relationship between frame rate and computational cost is readily quantified, how frame rates affect semantic information loss remains underexplored. Therefore, there is a clear need to systematically examine how different frame rates shape the quality of speech tokens, especially across languages with varying phonetic and acoustic characteristics.

In this paper, we examine the impact of frame rates on speech tokenizers through a language-specific lens, using Mandarin and English as contrasting examples. Our investigation begins with an analysis of ASR performance across various frame rates and an exploration of how linguistic identity, particularly tonal characteristics of Mandarin, affects codebook usage. Building on these insights, we propose strategies to mitigate semantic information loss in low-frame-rate speech tokenization, underscoring the value of considering language-specific properties in tokenizer design.

\section{Speech Tokenizer}
\begin{figure}[t]
  \centering
  \includegraphics[width=\linewidth]{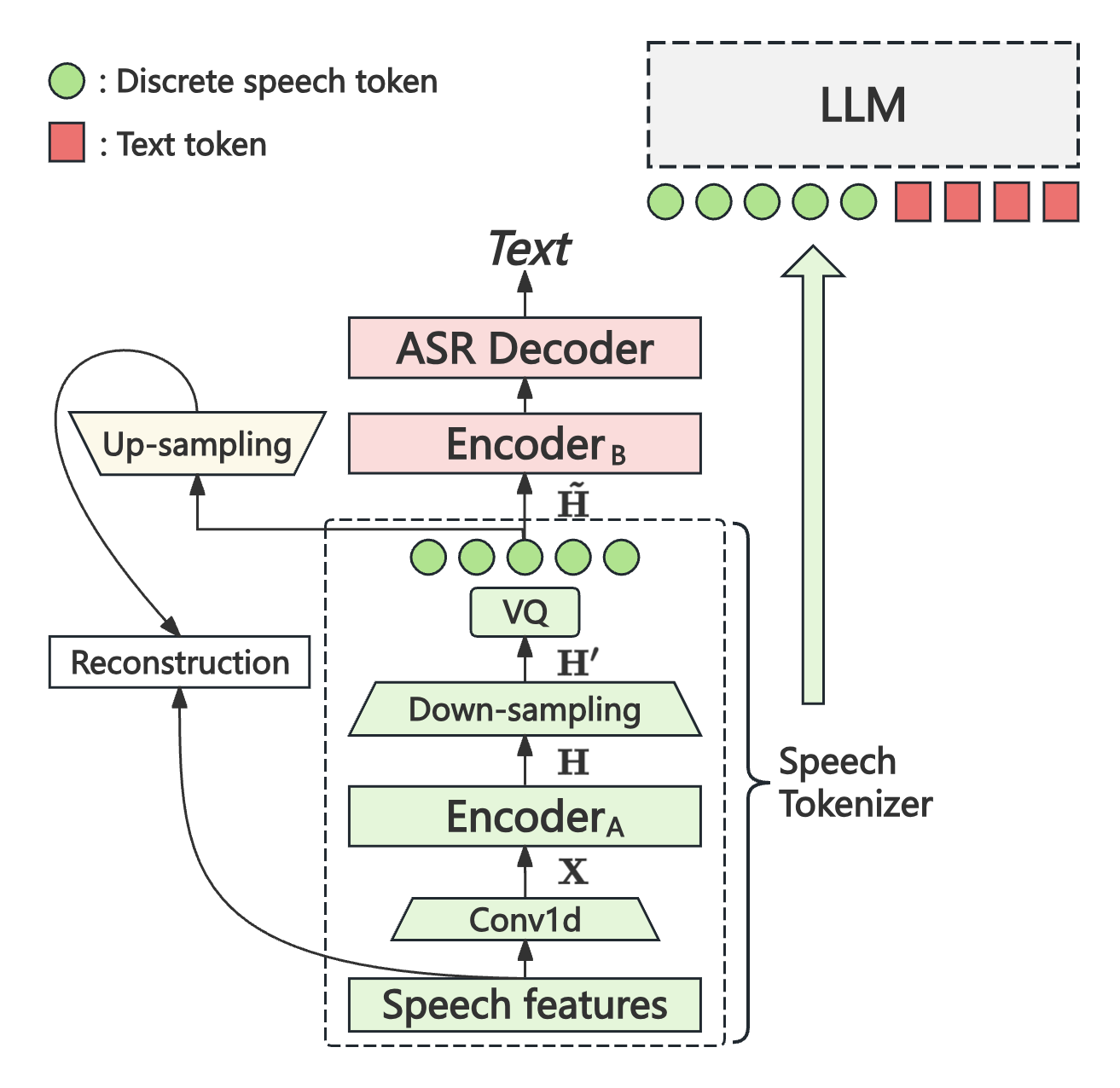}
  \caption{Speech tokenizer within a Transformer encoder-decoder ASR model and the downstream LLM.}
  \label{fig:speech_tokenizer}
\end{figure}

As shown in Fig.~\ref{fig:speech_tokenizer}, we adopt a similar paradigm with Cosyvoice to develop the speech tokenizer~\cite{du2024cosyvoice}, where the speech tokenizer is incorporated in a Transformer encoder-decoder ASR system. 

Given a speech signal, similar to the Whisper models, the speech signal is transformed into its log-mel spectrogram before being fed into a convolutional feature extractor. We define the output of the feature extractor as $\mathbf{X}=(\mathbf{x}_{t} \in \mathbb{R}^{F}| t=1, \ldots, T)$ and the paired token sequence as $W=(w_n \in \mathcal{V} | n=1, \ldots, N)$, where $\mathcal{V}$ is the vocabulary, $T$ and $N$ are the lengths of feature and token sequences, respectively, and $F$ is the dimension of the acoustic feature. The hidden representations $\mathbf{H}=(\mathbf{h}_{t} \in \mathbb{R}^{D}| t=1, \ldots , T)$ are then generated by $\mathrm{Encoder_{A}}$ from $\mathbf{X}$. Hidden representations $\mathbf{H}$ are next fed into a Vector Quantized-Variational AutoEncoder (VQ-VAE) module~\cite{vqvae}. Here, the VQ-VAE module comprises up-sampling and down-sampling convolutional neural networks (CNN) and a vector quantizer (VQ). The conv1d feature extractor, $\mathrm{Encoder_{A}}$, the down-sampling module, and VQ form a speech tokenizer, which is employed to tokenize the speech signals for LLMs. The down-sampling CNN reduces the sequence length of $\mathbf{H}$ and generates $\mathbf{H}^{\prime}=(\mathbf{h}^{\prime}_{t} \in \mathbb{R}^{D}| t=1, \ldots , T^{\prime})$, where each $\mathbf{h}^{\prime}_{t}$ is tokenized into a index $c_{k}$ of its nearest embedding in the codebook $C$ via 
\begin{equation}
  c_{t} = \mathrm{VQ}(\mathbf{h}^{\prime}_{t}, C) = \mathop{\mathrm{arg\,min}}\limits_{\mathbf{c}_n \in C} \vert\vert \mathbf{h}^{\prime}_{t} - \mathbf{c}_{n} \lvert\lvert
\label{eq:quantize}
\end{equation}
where $ \mathrm{VQ}(\cdot, \cdot)$ denotes the process of VQ and $\mathbf{c}_{n}$ is the embedding corresponding to index $n$. The input of $\mathrm{Encoder_{B}}$ involves the embeddings of these indexes achieved by (\ref{eq:quantize}), being $\tilde{\mathbf{H}}=(\mathbf{c}_{t} \in \mathbb{R}^{D}| t=1, \ldots , T^{\prime})$.

The VQ-VAE module down-samples and quantizes $\mathbf{H}$ into $\tilde{\mathbf{H}}$, which is the input of $\mathrm{Encoder_{B}}$. The joint optimization of the VQ-VAE and Transformer ASR model is performed via an objective function as
\begin{equation}
  \mathcal{L}_{\mathrm{all}}=\alpha \mathcal{L}_{\mathrm{ctc}} + \left ( 1-\alpha  \right ) \mathcal{L}_{\mathrm{att}} + \beta \mathcal{L}_{\mathrm{rec}},
  \label{eq:loss_all}
\end{equation}
where $\mathcal{L}_{\mathrm{ctc}}$ is the connectionist temporal classification (CTC) loss~\cite{ctc}, $\mathcal{L}_{\mathrm{att}}$ is the encoder-decoder ASR loss, and $\mathcal{L}_{\mathrm{rec}}$ is the mean square error (MSE) loss computed between the log-magnitude Mel-spectrogram and the output of the upsampling module for reconstruction. Parameters $\alpha$ and $\beta$ denote the weight associated with ASR and VQ-VAE during training, respectively.

\section{Dataset and Experiment}
\subsection{Dataset}
The training data comprise 80,000-hour paired speech-text data, with 40,000 hours of Mandarin, 20,000 hours of English, 15,000 hours of Cantonese, and 5,000 hours of Japanese. The test data involve the AISHELL-2 test set and LibriSpeech test-clean set~\cite{aishell2, librispeech}.

\subsection{Experiment setup}
The Transformer-based ASR model is composed of a feature preprocessor, 32 encoder layers, 32 decoder layers, and a CTC layer. Each Transformer layer employs 20 attention heads with dimension $D=1,280$, and the inner feed-forward layer is 5,120-dimensional. Input speech signals are converted into 128-channel log-magnitude Mel-spectrograms using a 25-ms window and 10-ms stride (i.e., 100 Hz frame rate), and then fed into a conv1d feature extractor. The feature extractor adopts a similar architecture as Whisper, with the output dimension being 1280~\cite{whisper}.

The VQ-VAE module consists of a down-sampling component, an up-sampling component, and a VQ module. The down-sampling component comprises three convolutional layers with kernel sizes of (3,3,5) and strides of (2,2,3), reducing the sequence length to the target frame rate. The up-sampling component restores the sequence length to the scale of the speech features with four residual blocks, each containing convolutional kernels of sizes (5,3,3,3) and strides of (3,2,2,2). The output of the up-sampling module has a frame rate of 100~Hz, being the reconstructed speech features. The VQ codebook contains 4096 entries.

The ASR model is first pre-trained on the 80,000-hour data (i.e., Our ASR in Table~\ref{tab:overall}). We then inserted the VQ-VAE module between the sixth and seventh layers of the Transformer encoder and fine-tuned the integrated system for 2 epochs using Adam optimizer with a batch size of 26~\cite{stepaudio}. The learning rate warms up from 0 to $2\times10^{-5}$ over 12,000 steps before decaying linearly, with $\alpha=0.3$ and $\beta=1.0$.

Force alignment was performed with a pre-trained Whisper.large model~\cite{liu2024aligning}. ASR performance is evaluated using word error rate (WER), with Mandarin ASR performance assessed at the character level.

\begin{table}[t]
\renewcommand{\arraystretch}{1.1}
\setlength{\tabcolsep}{2mm}{
\begin{tabular}{cccccc}
\toprule
\multirow{2}{*}{\textbf{Model}} & \textbf{Frame} & \multicolumn{2}{c}{\textbf{Codebook usage}}  & \multicolumn{2}{c}{\textbf{WER}} \\ 
     &  \textbf{Rate}   & AISHELL & LS-clean & AISHELL & LS-clean \\ \midrule
Whisper      & 50 Hz    & -   & -       & 4.87     & 2.15      \\ \midrule
Our ASR      & 50 Hz    & -   & -       & 2.85     & 1.92      \\ \midrule
S1           & 12.5 Hz  & 88.06      & 99.73    & 6.85      & 3.54                  \\ 
S2           & 8.33 Hz  & 78.54      & 98.39    & 9.38      & 5.80                  \\
S3           & 6.25 Hz  & 68.43      & 98.27    & 12.66      & 6.55                  \\ 
S4           & 5.00 Hz  & 67.77      & 97.53    & 27.20      & 8.48                  \\ \bottomrule
\end{tabular}}
\caption{Comparison of systems trained with various frame rates in terms of codebook usage (\%) and WER (\%) for AISHELL-2 test set and LibriSpeech test-clean set}\label{tab:overall}
\end{table}

\section{Results and Analysis}
\subsection{Overall comparison of various frame rates}
\label{sec:overall_fr}

\begin{figure*}[t]
  \centering
  \includegraphics[width=\linewidth]{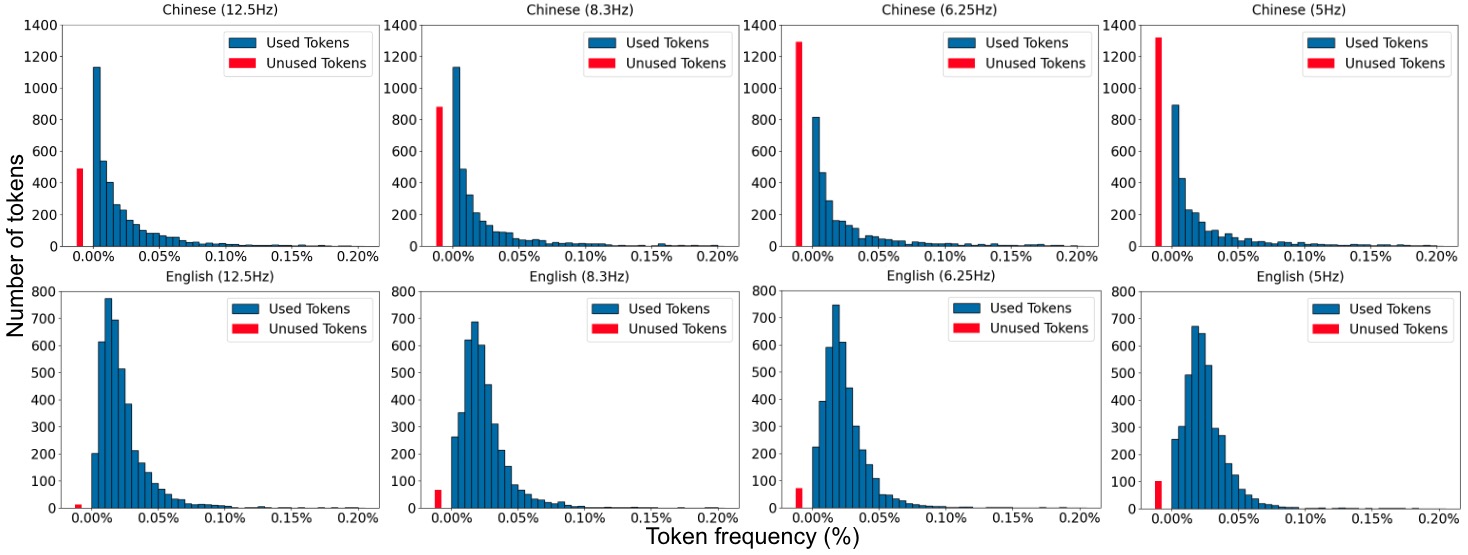}
  \caption{Token distribution under various frame rates from 12.5~Hz (left) to 5~Hz (right) for LibriSpeech test-clean set (bottom) and AISHELL-2 test set (top).}
  \label{fig:token_vs_freq}
\end{figure*}

We first compare our models and present the results in Table ~\ref{tab:overall}. Systems S1 through S4 incorporate the VQ-VAE module into our pre-trained ASR framework, each using different frame rates for the speech tokenizer. 

\begin{figure}[t]
  \centering
  \includegraphics[width=\linewidth]{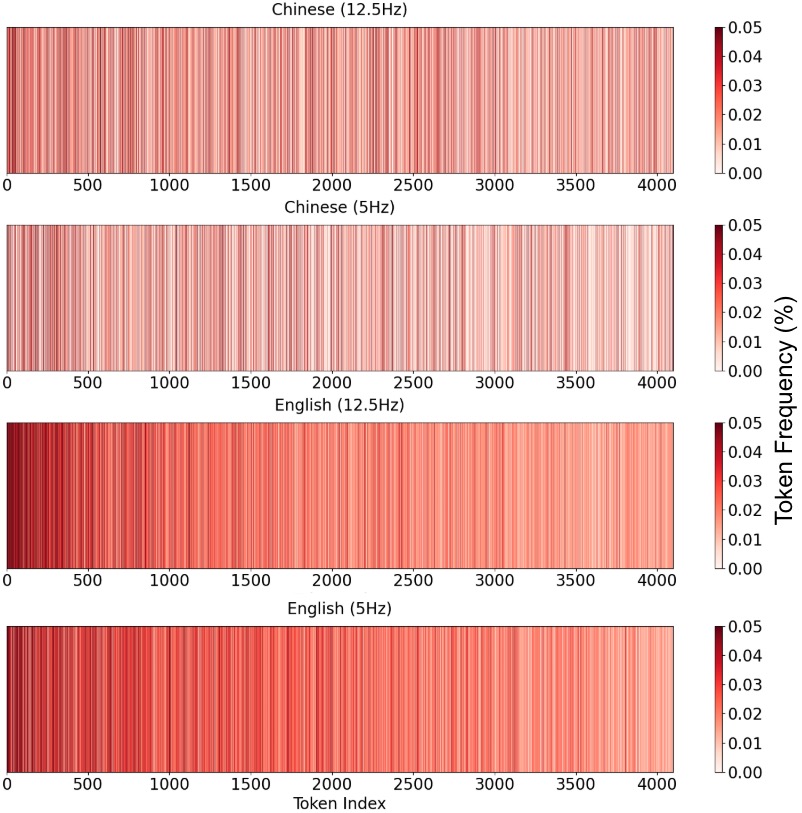}
  \caption{Visualization of codebook usage for Mandarin (top) and English (bottom) with 12.5~Hz and 5~Hz frame rates.}
  \label{fig:code_usage}
\end{figure}

It is not surprising that lower frame rates lead to lower ASR performance and codebook usage. This is because more aggressive compression entails a greater loss of acoustic information. Additionally, models suffer from a greater degradation in ASR performance and codebook usage for Mandarin data than for English data, though the Mandarin training data is twice as extensive as the English training data. Especially when using a frame rate of 5~Hz, S4 exhibits significantly degraded ASR performance on Mandarin data with a WER of 27.20\%, whereas the English ASR performance remains within acceptable bound. We also compute statistics on codebook usage and present the results in Figures~\ref{fig:token_vs_freq} and \ref{fig:code_usage}. Our analysis indicates that decreasing the frame rate results in a higher proportion of unused tokens in the Mandarin dataset compared to the English dataset. 

The above could result from the distinct linguistic characteristics of these two languages. Specifically, Mandarin is a tonal language in which tone is integral to each syllable's meaning\textemdash even slight pitch variations can differentiate entirely distinct words. In contrast, English is non-tonal, relying on stress and intonation patterns at the phrase or word level rather than on syllable- or mora-level pitch changes for lexical meaning~\cite{li2013spoken, liu22e_interspeech}. Moreover, Mandarin characters typically have shorter durations than English words. Consequently, speech tokens extracted at a low frame rate such as 5~Hz, with each token spanning approximately 200~ms, may capture the primary information contained in a Mandarin character while failing to fully represent an English word. This discrepancy poses a challenge for tokenizing Mandarin speech, as the tokenizer may inadvertently separate the tonal component from the segmental (non-tonal) content of a syllable, resulting in significant information loss in both aspects. As opposed to Mandarin, English is non-tonal and relies on stress and intonation patterns that span longer temporal windows. Therefore, the low frame rate affects it less severely.

\begin{CJK}{UTF8}{gbsn}
\begin{figure}[t]
  \centering
  \includegraphics[width=\linewidth]{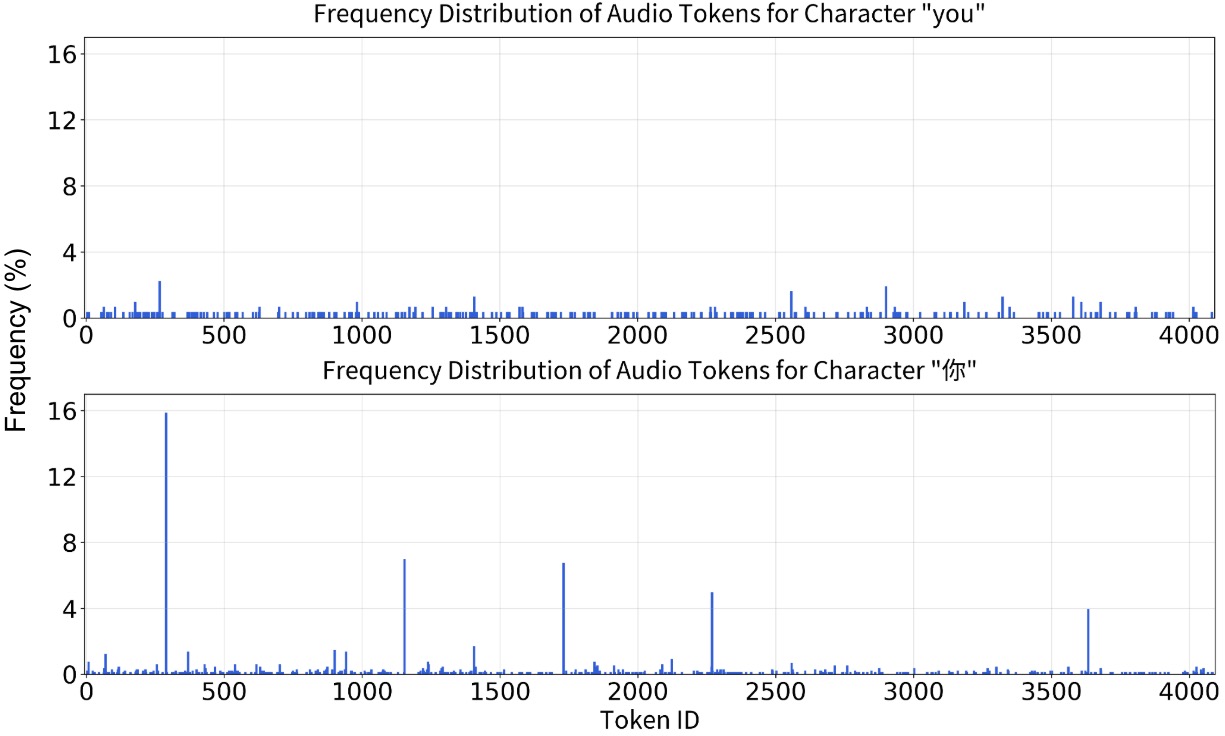}
  \caption{Comparison of how the speech for a Mandarin character ``你" (bottom) and its English equivalent (top) ``you'' map to each codebook entry based on usage frequency.}
  \label{fig:you_token_vs_freq}
\end{figure}
\end{CJK}
\begin{CJK}{UTF8}{gbsn}
\footnotesize
\begin{table}[h!]
\renewcommand{\arraystretch}{1.2}
\setlength{\tabcolsep}{0.3mm}{
{\footnotesize
\begin{tabular}{c|c|l}
\toprule
\multicolumn{1}{c}{\textbf{Padding}} & \multicolumn{1}{c}{\textbf{Mandarin ASR}} & \multicolumn{1}{c}{\textbf{English ASR}} \\ \midrule
0.00  & 你该重新\textcolor{red}{指点小鬼}了 & Tomorrow \textcolor{red}{we'll see examinations} \\ 
0.01  & 你该重新\textcolor{red}{指点小鬼}了 & Tomorrow \textcolor{red}{i will see that there may be some} \\ 
0.02  & 你该重新\textcolor{blue}{制定}\textcolor{red}{法规}了 & Tomorrow \textcolor{red}{we'll see if that may be the same} \\ 
0.03  & 你该\textcolor{red}{从先知顶上跪}了 & Tomorrow \textcolor{red}{we'll see the family} \\ 
0.04  & 你该\textcolor{red}{从先知顶上跪}了 & Tomorrow \textcolor{blue}{is the} examination \\ 
0.05  & 你该重新\textcolor{red}{指定上位}了 & Tomorrow \textcolor{red}{i will see if they have an answer} \\ 
0.06  & 你该重新\textcolor{red}{指定上位}了 & Tomorrow \textcolor{red}{i will see you generally} \\
0.07  & 你该重新\textcolor{red}{指定上位}了 & \textcolor{red}{To my request} \\ 
0.08  & 你该重新\textcolor{blue}{制定}\textcolor{red}{社会}了 & Tomorrow \textcolor{red}{i will seek some relation} \\ 
0.09  & 你该重新\textcolor{blue}{制定}\textcolor{red}{社会}了 & Tomorrow \textcolor{red}{i will seek them in person} \\ 
0.10  & 你该重新\textcolor{red}{指定上回}了 & Tomorrow \textcolor{red}{i will seek some relation} \\ 
0.11  & 你该重新\textcolor{red}{决定上回}了 & Tomorrow \textcolor{red}{we'll see if sam has anything} \\ 
0.12  & 你该重新\textcolor{red}{决定上回}了 & Tomorrow \textcolor{red}{i will see you again} \\ 
0.13  & 你该重新\textcolor{blue}{制定校规}了 & Tomorrow \textcolor{red}{i will see you again} \\
0.14  & 你该重新\textcolor{blue}{制定校规}了 & Tomorrow \textcolor{red}{we'll see you} \\ 
0.15  & 你该重新\textcolor{blue}{制定校规}了 & Tomorrow \textcolor{red}{i will see you generally} \\ 
0.16  & 你该重新\textcolor{blue}{制定}\textcolor{red}{小}\textcolor{blue}{规}了 & Tomorrow \textcolor{red}{i will see a gentleman} \\
0.17  & 你该重新\textcolor{blue}{制定}\textcolor{red}{小}\textcolor{blue}{规}了 & Tomorrow \textcolor{red}{we'll see them} \\ 
0.18  & 你该重新\textcolor{blue}{制定}\textcolor{red}{小}\textcolor{blue}{规}了 & \textcolor{red}{Tomorrow} \\ 
0.19  & 你该重新\textcolor{red}{指定上官}了 & \textcolor{red}{To morrow} \\ 
0.20  & 你该重新\textcolor{red}{指点小鬼}了 & \textcolor{red}{To morrow} \\ 
0.21  & 你该重新\textcolor{red}{指点小鬼}了 & \textcolor{red}{To morrow} \\ 
0.22  & 你该重新\textcolor{blue}{制定}\textcolor{red}{法规}了 & Tomorrow \textcolor{red}{we'll see them} \\ 
0.23  & 你该重新\textcolor{red}{指定上位}了 & Tomorrow \textcolor{red}{we'll see the family} \\ 
0.24  & 你该重新\textcolor{red}{指定上官}了 & Tomorrow \textcolor{red}{i will see that i may listen} \\ 
0.25  & 你该重新\textcolor{red}{指定上会}了 & Tomorrow \textcolor{red}{we see them in the sun} \\ 
0.26  & 你该重新\textcolor{red}{指定上会}了 & Tomorrow \textcolor{red}{we see them} \\ 
0.27  & 你该重新\textcolor{red}{指定上会}了 & Tomorrow \textcolor{red}{we see them listen} \\ 
0.28  & 你该重新\textcolor{blue}{制定}\textcolor{red}{商会}了 & Tomorrow \textcolor{red}{i will see him} \\ 
0.29  & 你该重新\textcolor{blue}{制定}\textcolor{red}{商会}了 & Tomorrow \textcolor{red}{i will seek some medicine} \\ 
0.30  & 你该重新\textcolor{red}{决定上回}了 & \textcolor{red}{Come on, we'll see} \\ 
0.31  & 你该重新\textcolor{red}{指定上回}了 & \textcolor{red}{To morrow} \\ 
0.32  & 你该重新\textcolor{red}{决定上回}了 & \textcolor{red}{To morrow} \\ 
0.33  & 你该重新\textcolor{blue}{制定校规}了 & Tomorrow \textcolor{red}{we'll see them} \\ 
0.34  & 你该重新\textcolor{blue}{制定校规}了 & Tomorrow \textcolor{red}{i will see them} \\  
0.35  & 你该重新\textcolor{blue}{制定}\textcolor{red}{小}\textcolor{blue}{规}了 & Tomorrow \textcolor{red}{i will see them} \\ 
0.36  & 你该重新\textcolor{blue}{制定}\textcolor{red}{小}\textcolor{blue}{规}了 & Tomorrow \textcolor{red}{we'll see examinations} \\ 
0.37  & 你该重新\textcolor{red}{指定上官}了 & \textcolor{red}{To morrow} \\ 
0.38  & 你该重新\textcolor{red}{指定上官}了 & \textcolor{red}{To morrow} \\ 
0.39  & 你该重新\textcolor{blue}{制定}\textcolor{red}{小}\textcolor{blue}{规}了 & \textcolor{red}{To morrow} \\ 
0.40  & 你该重新\textcolor{blue}{制定}\textcolor{red}{小}\textcolor{blue}{规}了 & \textcolor{red}{To morrow} \\ \bottomrule
\end{tabular}}}
\caption{Impact of various padding durations (in seconds) on Mandarin and English ASR outputs using 5Hz speech tokenizer. Ground truth: Mandarin - ``你该重新制定校规了'', English - ``Tomorrow is the examination''. Errors and correctly recognized characters are marked in red and blue, respectively}\label{tab:padding}
\end{table}
\end{CJK}
\subsection{Impact of linguistic identity}
\begin{CJK}{UTF8}{gbsn}
To further validate our statement in Sec.~\ref{sec:overall_fr} that a low-frame-rate speech tokenizer may capture primary information of a Mandarin character more effectively than an English word, we analyze a commonly used Mandarin character ``你" and its English equivalent ``you" as examples. 

We summarize the embeddings within the VQ codebook mapped to the acoustic patterns for ``你" and ``you" during inference, respectively. For each target character and word, we perform forced alignment to obtain the start and end timestamps of the corresponding speech segment. Next, utterances from the English and Mandarin test sets are tokenized using the speech tokenizer from the pre-trained S4 system (see Table~\ref{tab:overall}), and the codebook entries associated with the target segments are recorded. Finally, we compute the frequency of occurrence for each codebook entry by dividing the number of times it is mapped to the target segment by the total number of mappings.

The statistics in Figure~\ref{fig:you_token_vs_freq} show that the speech patterns of both the Chinese character ``你" and the English word ``you" are typically associated with a fixed set of high-frequency tokens in the codebook. However, ``你" is consistently mapped to the same token more often than ``you." This suggests that speech tokens for a Mandarin character tend to capture more uniform patterns across different utterances compared to those for an English word. Consequently, this results in a greater reduction in codebook usage for Mandarin compared to English, as demonstrated in Figures~\ref{fig:token_vs_freq} and \ref{fig:code_usage}, further supporting our claim in Sec.~\ref{sec:overall_fr}.
\end{CJK}

\begin{figure*}[h!]
  \centering
  \includegraphics[width=0.85\linewidth]{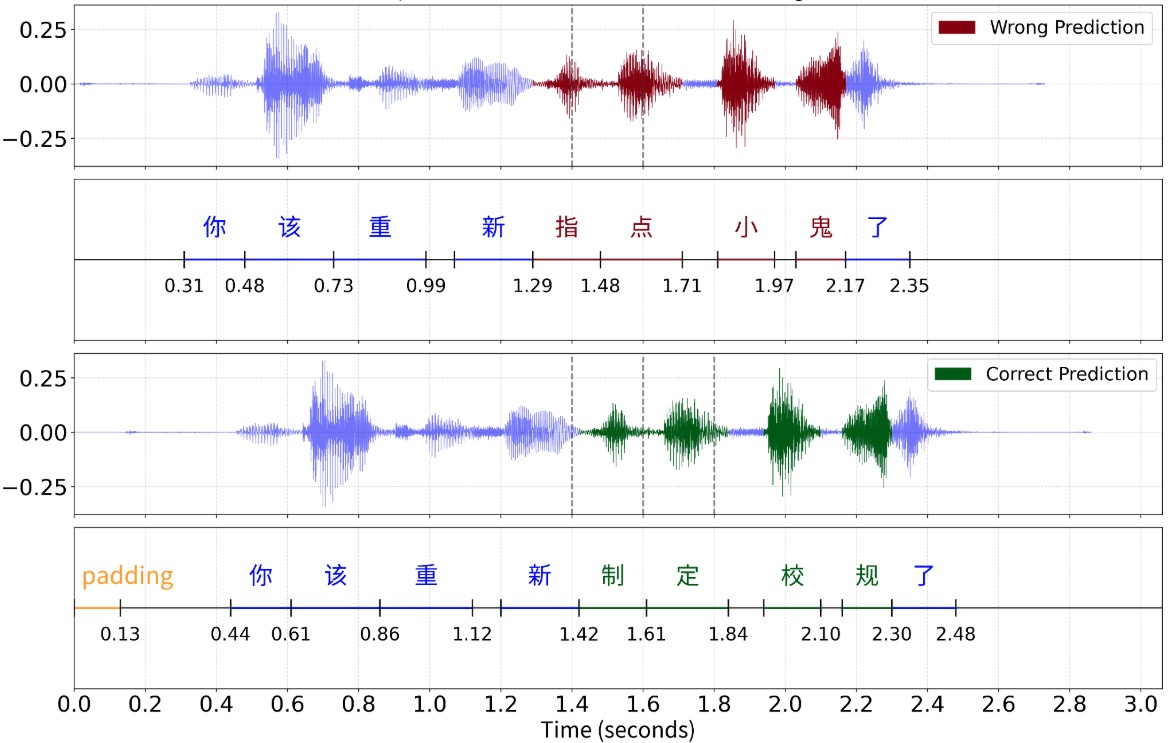}
  \caption{Comparison of the ASR performance of a Mandarin case before (top) and after (bottom) padding. Errors, characters corrected by padding, and the padding are marked in red, green, and orange, respectively.}
  \label{fig:audio_map}
\end{figure*}

\subsection{Padding to retain speech information}
To explore and mitigate the impact of the linguistic characteristics of Mandarin on low-frame-rate speech tokenizers, we propose padding the speech signal to prevent the inadvertent separation of the tonal component of a Mandarin character from its segmental (non-tonal) content.

As a case study, we examine one Mandarin utterance and one English utterance, padding each audio sample with silence segments of varying durations. These silence segments, which are extracted from the respective examples, are inserted at the beginning of each sample. By shifting the waveform via padding, the speech tokenizer can capture the primary information of each Mandarin character more effectively.

Table~\ref{tab:padding} presents the ASR outputs generated by system S4 for the original and padded speech signals. We observe that different padding durations yield varying ASR performance. In particular, system S4 demonstrates optimal performance for the Mandarin utterance when padding durations are approximately 0.13~s and 0.33~s, while higher error rates are observed when the padding deviates from these values. This suggests that applying appropriate padding to Mandarin utterances can effectively realign the tonal component with its corresponding segmental content, thereby preserving complete information. As opposed to the observation regarding the Mandarin example, padding does not necessarily improve English ASR performance, although system S4 may benefit from a specific padding value in this particular case.

In addition, we visualize both the original waveform and the waveform padded with an optimal value (i.e., 0.13~s). As shown in Figure~\ref{fig:audio_map}, the speech tokenizer cuts off portions of the Mandarin characters, potentially separating the tonal component from the non-tonal content and leading to ASR errors. This visualization highlights the impact of the linguistic characteristics of Mandarin on low-frame-rate speech tokenizers. By applying optimal padding, the waveform is shifted such that fewer Mandarin characters are truncated. This allows the tokenizer to retain a more complete speech signal, thereby improving ASR performance.

Although developing a padding strategy for large-scale data may not be practical, it offers a promising avenue for designing a multilingual speech tokenizer. Since multilingual TTS or speech dialogue systems often involve languages with significantly different linguistic identities (e.g., tonal versus non-tonal), an adaptive technique that is applicable across all languages is preferable.

\section{Conclusion}
\label{sec:conclusion}
In this work, we investigated the impact of frame rates on speech tokenizers with a case study on English and Mandarin, typical non-tonal language and tonal language. Experiments indicate that low-frame-rate speech tokenizers face greater information loss challenges in tonal languages compared to non-tonal languages. Moreover, this information loss can be mitigated by padding the speech signals, which prevents the tokenizer from inadvertently separating the tonal component from the non-tonal content of a Mandarin character. These findings underscore the importance of tailoring speech tokenization strategies to the specific linguistic characteristics of the target language. While our study is limited in language scope, it provides valuable insights and a foundation for future research on adaptive padding strategies and their integration into speech tokenizers.

\section*{Acknowledgment}
The authors would like to thank the College of Computing and Data Science (CCDS) at Nanyang Technological University, Singapore, and StepFun for their support.


\begin{thebibliography}{00}
\bibitem{vall_e} W. Wang et al., ``Neural codec language models are zero-shot text to speech synthesizers,'' arXiv preprint arXiv:2301.02111, 2023.
\bibitem{du2024cosyvoice} Y. Du et al., ``CosyVoice: A Scalable Multilingual Zero-shot Text-to-speech Synthesizer based on Supervised Semantic Tokens,'' arXiv preprint arXiv:2407.05407, 2024.
\bibitem{chu2023qwen} Y. Chu et al., ``Qwen-Audio: Advancing Universal Audio Understanding via Unified Large-Scale Audio-Language Pretraining,'' arXiv preprint arXiv:2311.07919, 2023.
\bibitem{defossez2024moshi} A. Defossez et al., ``Moshi: a speech-text foundation model for real-time dialogue,'' arXiv preprint arXiv:2410.00037, 2024.
\bibitem{zeng2024glm} A. Zeng et al., ``GLM-4-Voice: Towards Intelligent and Human-Like End-to-End Spoken Chatbot,'' arXiv preprint arXiv:2410.12608, 2024.
\bibitem{gense25} GenSE Team, ``GenSE: A Series of Audio Generation Models,'' arXiv preprint arXiv:2501.03046, 2025.
\bibitem{chen2022beats} S. Chen et al., ``BEATs: Audio Pre-Training with Acoustic Tokenizers,'' arXiv preprint arXiv:2212.09058, 2022.
\bibitem{zhang2024speechtokenizer} X. Zhang et al., ``SpeechTokenizer: Unified Speech Tokenizer for Speech Large Language Models,'' arXiv preprint arXiv:2308.16692, 2024.
\bibitem{ji2024wavtokenizer} Y. Ji et al., ``WavTokenizer: an Efficient Acoustic Discrete Codec Tokenizer for Audio Language Modeling,'' arXiv preprint arXiv:2408.16532, 2024.
\bibitem{zhang2025facespeak} H. Zhang et al., ``FaceSpeak: Towards Multi-Modal Speech Generation from Visual Input,'' arXiv preprint arXiv:2501.02094, 2025.
\bibitem{encodec} A. Défossez et al., ``High fidelity neural audio compression,'' arXiv preprint arXiv:2210.13438, 2022.
\bibitem{soundstream} N. Zeghidour et al., ``SoundStream: An End-to-End Neural Audio Codec,'' IEEE/ACM Transactions on Audio, Speech, and Language Processing, vol. 30, pp. 495-507, 2022.
\bibitem{transformer} A. Vaswani et al., ``Attention is all you need,'' Advances in neural information processing systems, vol. 30, 2017.
\bibitem{vqvae} A. Van Den Oord et al., ``Neural discrete representation learning,'' Advances in neural information processing systems, vol. 30, 2017.
\bibitem{ctc} A. Graves et al., ``Connectionist temporal classification: labelling unsegmented sequence data with recurrent neural networks,'' in Proceedings of the 23rd international conference on Machine learning, 2006, pp. 369-376.
\bibitem{aishell2} J. Du et al., ``AISHELL-2: Transforming Mandarin ASR Research Into Industrial Scale,'' arXiv preprint arXiv:1808.10583, 2018.
\bibitem{librispeech} V. Panayotov et al., ``Librispeech: an ASR corpus based on public domain audio books,'' in 2015 IEEE international conference on acoustics, speech and signal processing (ICASSP), 2015, pp. 5206-5210.
\bibitem{whisper} A. Radford et al., ``Robust speech recognition via large-scale weak supervision,'' in International Conference on Machine Learning, 2023, pp. 28492-28518.
\bibitem{stepaudio} H. Zhang et al., ``StepAudio: A framework for pre-trained audio models training and reasoning,'' 2024. [Online]. Available: https://github.com/StepFun-AI
\bibitem{liu2024aligning} H. Liu et al., ``Aligning Speech with Large Language Models for Automatic Speech Recognition,'' arXiv preprint arXiv:2404.10543, 2024.
\bibitem{li2013spoken} K. Li and M. Akagi, ``A three-layered model for expressive speech perception,'' Speech Communication, vol. 55, no. 4, pp. 530-540, 2013.
\bibitem{liu22e_interspeech} H. Liu et al., ``Tone recognition in Mandarin Chinese using convolutional neural networks,'' in Proc. Interspeech 2022, 2022, pp. 1876-1880.
\end{thebibliography}
\end{document}